Pressed to Migrate:

News Sentiment as a Predictor for American Domestic Migration

**Benjamin Lane**\*, Simeon Sayer

benlane515@gmail.com, ssayer@fas.harvard.edu

\*Primary Author

**ABSTRACT**

This paper goes into depth on the effect that US News Sentiment from national newspapers has on US interstate migration trends. Through harnessing data from the New York Times between 2010 and 2020, an average sentiment score was calculated, allowing for data to be entered into a neural network. Then a logistic regression model was used to predict interstate migration. The results indicate the model was highly accurate as the mean margin of error was +/- 900 citizens. The predictions from the model were compared with the US Census data from 2010 to 2020 that was used to train the model. Since the input for the model was not exposed to any migration data, the model clearly demonstrated that its results were drawn from sentiment data alone. These findings are significant as they indicate that the role of the press could be used as a predictor for domestic migration which can help the government and businesses understand better what is influencing people to move to certain places.

**II. INTRODUCTION**

People move for a variety of reasons, whether that be for a new job, to be closer to family, or for better access to schools. While there are many factors at play in any individual move, this paper aims to investigate whether press coverage of different locations can act as a leading indicator or a proxy for these moves. By exploring the role of the press, particularly the sentiment of articles written about a place and the difference in press coverage between any two states, a machine learning model can be built to predict internal migration, and explore the extent to which the relationship works.

Using U.S. Census data that monitors and tracks how many people move between states and a New York Times dataset, a complete training set was created for the sentiments and migrations from 2010 to 2020. Using this dataset, a neural network model was built to predict the rate at which migration will occur, calculated according to the difference in sentiment scores for those two states in a given year.

This project develops a predictive model that can be used to anticipate migration patterns. Additionally, the experiment gives insight into the role of the press as a predictor for human behavior.

## III. BACKGROUND

The idea of using press sentiment as a predictor for market trends and human decisions has been explored before. For instance, Hausler *et al*'s 2018 paper [1] aimed to use press reports that contained the key words 'real estate' to predict US house prices, which was successful, finding an $R^2$ value of 0.82. Similar work based on twitter sentiment for stock prices has been well documented [2].

There have also been attempts to predict the internal migration rates using other data points. Lin *et al* aimed to predict internal migration patterns using Bing search results, as provided by Microsoft Research [3]. They were able to generate an $R^2$ coefficient correlation of 0.72, based on the prevalence of a query in a certain state and the rate at which there was movement in and out of that state. There have also been attempts to use other proxies, such as entrepreneurial levels to predict internal movement, with studies finding that more entrepreneurial states are more attractive to young people, who are in turn more likely to move.

However, this comes with a fundamental assumption that people are actively aware and researching entrepreneurial states. In reality, this paper hypothesizes it does not matter what the true statistics around a state are, what is more important is how those states are covered in the press. That is, the ultimate predictor for where people want to move to is what people think about different states, which is informed by social factors, such as word of mouth, or by the coverage of those states in the press.

## IV. DATASET

In order to build a model that was able to draw a prediction of the inter-state migration levels from the press coverage of those different states, each side of the data needs to be represented—the press and the migration.

The migration level data comes from the US census bureau [4] which collects data between all combinations of states via survey with a large sample size. This data is readily available for the years in the range of interest, between 2010 and 2020. A margin of error for each value is also provided, to contextualize the accuracy of a model.

Press sentiment data was calculated from a repository of New York Times articles, containing every article published since 2000 [5]. This contains the metadata, publication date, and the key words of all articles published. This choice for the source of data was easy to make. The New York Times has been widely popular since long before 2000, so the articles are likely to have a significant impact on readers. The paper is national and covers news from all over the country, making the data very valuable. The news covers a wide range of topics, meaning the many different factors that could be influencing people to migrate within the United States are represented. The New York Times is also able to maintain both a national view and a local view on many of the objects. However, a potential issue was overlap between states' names and words with other meanings. For example, many articles that would be classified as related to Washington State would actually be referring to Washington D.C. These were deleted in the python code.

**V. Methodology/Models**

This project used the Python coding language in Google Colaboratory to work upon the datasets and create the models. The Python functions used created a unique dataset for all 10 years for each state, depending on whether the state's name is included in the article. Next, the average sentiment score was determined, between -1 and 1, for all the articles in each of the years from 2010-2019. This was calculated using a pre-trained sentiment analyzer called VADER[6], which has been trained on social media data, which applies well to that of press reportings, as there is much duplication between the two datasets. For instance, a synopsis of the news is likely to appear within a tweet. VADER is also well suited to this task as it reliably produces a score of zero for neutral text, which is beneficial as not all news stories will express a sentiment, such as a weather report. By applying the VADER sentiment analysis to every news article published by the NYT over the 10 year period in question, and identifying news articles that refer to a specific state, 500 data points are generated. When state sentiment data is matched to the corresponding immigration rates of that state for that year, a full dataset is created, that is then used to train and test the model.

With the Sklearn library, 80 percent of the data was used to train the model, while the remaining 20 percent was used for testing. This adheres to the industry standard for machine learning models, and was found to provide reliable results in that it gave enough training data for the model to extract trends, but also enough testing data to verify the validity of the model.

First, the study examined a logistic regression of the relationship between sentiment scores and migration flows. This provided interesting trends, but with low accuracy, indicating a higher degree of complexity was needed to sufficiently train the model. Hence, the next model trained was a 4-layer neural network with an input and output size of 50, representing one value for each state. This was constructed using the TensorFlow library, and a visualization of the structure can be found in Figure 1 (below). The model's input had no access to any migration data, allowing to clearly be able to see if the model was able to draw conclusions from sentiment data alone. The model's performance on the unseen testing data is of the same of magnitude as the opposed to the training data it used to find its conclusions, and increased accuracy in one induced a increase in performance for the other, indicating that while the model indeed performed better on data that it had already seen, it was not actively overfitting or memorizing the data in a way that hindered the model.

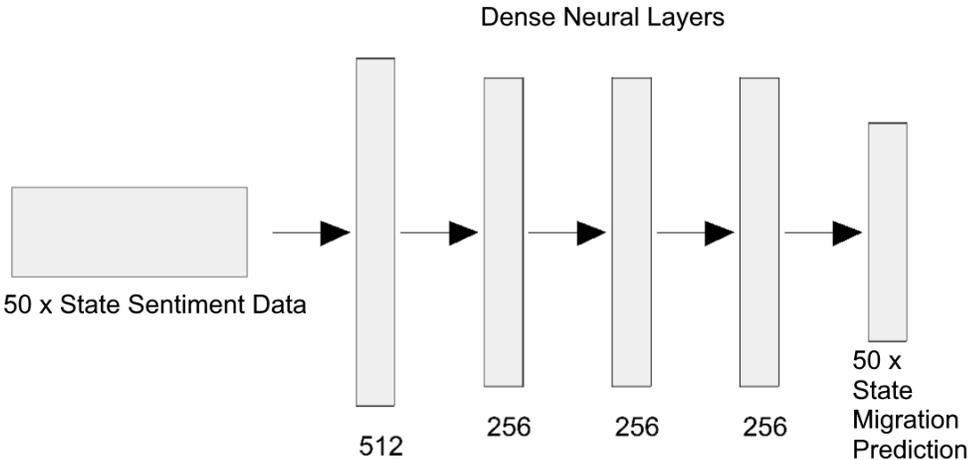

Fig. 1: Model Structure Visualization

**VI. Results and Discussion**

Through testing the model, this study determined that using news sentiment data alone, the model was able to predict the migration rates between any two states at an average (margin of error) level of either +900 or -900. This number clearly indicates that the model is strong and that sentiment is a useful predictor for the interstate migration rates. The impressiveness of this model is more clearly exhibited when the predictions and the actual values are compared side by side.

While 900 may seem like a significant amount, when contextualized with the populations of states, it becomes rather small. As shown, in most cases, +/- 900 allows for a very close approximation of migration data. In some cases, the model misses badly as a percentage of the overall migration out of that state, but these are almost always in states with a very low migration level, such that while the comparative error is high, the absolute difference compared to larger states is much lower. For instance, the model actually predicts an impossible negative value for Rhode Island, but the true value was so negligible, if anything this indicates the model is very confident that the migration rate will be extremely small. These very accurate predictions clearly indicate that newspaper sentiment analysis can be a meaningful predictor for interstate migration. Even if there were confounding factors that were overlooked in the research, the model proved to work very well, which is impressive and interesting.

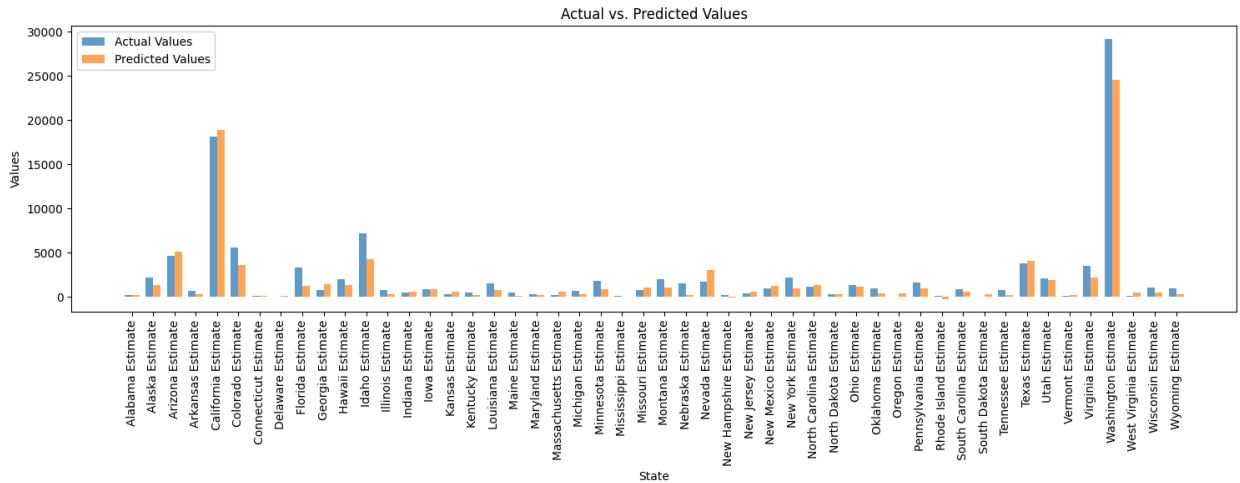

Fig. 2: Actual Values compared to Model Predictions for State Migration

The training of the model was also of interest. The initial loss of the model was very high, which decreased exponentially, as shown in the loss over epoch graph in Figure 3. The decision to allow the model to train for further epochs, despite diminishing returns, was that the absolute value of the loss was continuing to decrease in a linear fashion, indicating that further training beyond 2000 epochs may continue to improve results without substantial risk of overfitting.

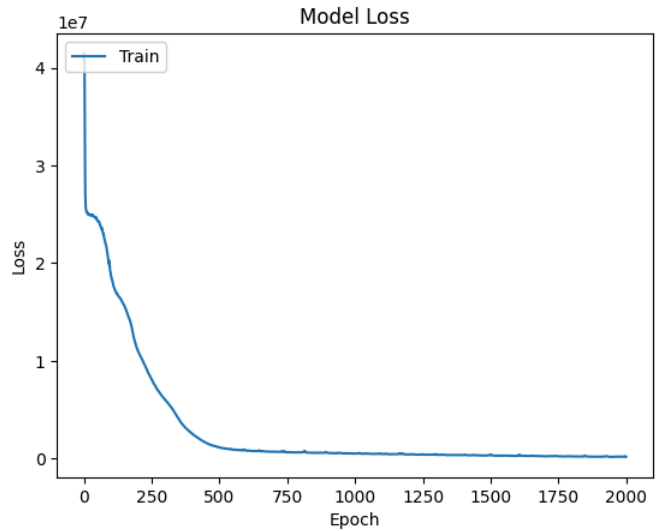

Fig. 3: Loss over Epochs

**VII. Conclusion**

This paper described the methods used to predict domestic U.S. migration. The data was taken from the U.S. Census and the internet archive and implemented into machine learning models in Python. First, sentiment scores were determined from the articles and then used those for a second model that incorporated the migration data. The data was split by the standard 80:20 to train and test the model and the results from the predictions indicated that the model works well. The datasets used were fairly simple and there was little room for error or hidden problems as long as the Python coding language was used correctly. This also ensured that the relationship was being tested in the predicted direction and that the sentiment wasn't changing due to the migration. Some possible ways to make it better are by using a broader variety of news or media sources in order to gain more data, broaden conclusions, and increase sentiment granularity. One limitation of this project is that it did not account for media sources other than newspapers. The revenue of newspaper publishers in the US dropped by 52% between 2002 and 2020, from $46

million to $22 million[7]. This indicates an overall reduced reliance on Newspapers among Americans. A future study could include social media data to get a better understanding of how people get their information today.

**REFERENCES**

[1] Hausler, J., Ruscheinsky, J., & Lang, M. (2018). News-based sentiment analysis in real estate: a machine learning approach. *Journal of Property Research*, *35*(4), 344–371. https://doi.org/10.1080/09599916.2018.1551923

[2] Ranco, G., Aleksovski, D., Caldarelli, G., Grcar, M., & Mozetic, I. (2015). The Effects of Twitter Sentiment on Stock Price Returns. *PLOS ONE*, *10*(9), e0–e0138441. https://doi.org/10.1371/journal.pone.0138441

[3] Lin, A. Y., Cranshaw, J., & Counts, S. (2019). Forecasting U.S. Domestic Migration Using Internet Search Queries. *The World Wide Web Conference*, 1061–1072. https://doi.org/10.1145/3308558.3313667

[4] U.S. Census Bureau. (n.d.).*State-to-State Migration Flows* . American Community Survey. Retrieved July, 2024, from https://www.census.gov/data/tables/time-series/demo/geographic-mobility/state-to-state-migration.html

[5] Kaggle Dataset Repository, *NYT Articles: 2.1M+ (2000-Present) Daily Updated*, Retrieved July, 2024, from https://www.kaggle.com/datasets/aryansingh0909/nyt-articles-21m-2000-present

[6] Hutto, C.J. & Gilbert, E.E. (2014). *VADER: A Parsimonious Rule-based Model for Sentiment Analysis of Social Media Text*. Eighth International Conference on Weblogs and Social Media (ICWSM-14). Ann Arbor, MI, June 2014.

[7] Redline. "US Newspapers Statistics." *Redline Digital*, https://redline.digital/us-newspapers-statistics/#:~:text=US%20newsroom%20employment%20fell%20by,while%2037%25%20used%20social%20media. Accessed 12 Jan. 2025.